\documentclass[10pt,twocolumn,letterpaper]{article}

 \usepackage{cvpr}              
\usepackage{graphicx}
\usepackage{amsmath}
\usepackage{amssymb}
\usepackage{booktabs}
\usepackage{bm}
\usepackage{multirow}
\usepackage{makecell}

\usepackage[pagebackref,breaklinks,colorlinks,citecolor=cvprblue]{hyperref}

\def\paperID{13873} 
\def\confName{CVPR}
\def\confYear{2024}

\title{A Single Simple Patch is All You Need for AI-generated Image Detection}

\author{Jiaxuan Chen\textsuperscript{\rm 1}, Jieteng Yao\textsuperscript{\rm 2}, Li Niu\textsuperscript{\rm 2}\thanks{Corresponding author.} \\
\textsuperscript{\rm 1} Nanjing University of Aeronautics and Astronautics\\
\textsuperscript{\rm 2} MoE Key Lab of Artificial Intelligence, Shanghai Jiao Tong University\\
{\tt \small chen\_jiaxuan@nuaa.edu.cn,\quad\{yjt18122666691,ustcnewly\}@sjtu.edu.cn}
}

\begin{document}
\maketitle
\definecolor{cvprblue}{rgb}{0.21,0.49,0.74}
\begin{abstract}
The recent development of generative models unleashes the potential of generating hyper-realistic fake images. To prevent the malicious usage of fake images, AI-generated image detection aims to distinguish fake images from real images. However, existing method suffer from severe performance drop when detecting images generated by unseen generators. We find that generative models tend to focus on generating the patches with rich textures to make the images more realistic while neglecting the hidden noise caused by camera capture present in simple patches. In this paper, we propose to exploit the noise pattern of a single simple patch to identify fake images. Furthermore, due to the performance decline when handling low-quality generated images, we introduce an enhancement module and a perception module to remove the interfering information. Extensive experiments demonstrate that our method can achieve state-of-the-art performance on public benchmarks. 
\end{abstract}

\section{Introduction} \label{sec:intro}
In recent years, the rapid advancement of generative models (\emph{e.g.}, GAN~\cite{goodfellow2014generative}, diffusion model~\cite{rombach2022high}) enables the creation of fake images. The generated images are so strikingly realistic that human observers can hardly distinguish them from real images. This raises the concern about malicious usage and negative social impacts. For example, realistic fake images could be used to spread false or misleading information with deceptive intentions. Considering the unprecedented power of generative models and their potential abuse, there is an increasing need to train a detector to automatically recognize AI-generated images, which is referred to as AI-generated image detection. 

However, this task remains challenging due to the poor generalization capability of learning-based detector. Early works~\cite{cozzolino2018forensictransfer,zhang2019detecting,wang1909fakespotter}  found that the detector trained on images generated by one GAN architecture performs poorly when tested on others. To overcome this challenge, various detection methods have been developed, including spatial domain methods~\cite{durall2020watch,jeong2022bihpf,nataraj2019detecting} and frequency domain methods~\cite{jeong2022frepgan,frank2020leveraging,qian2020thinking}. Some recent methods~\cite{wang2020cnn,chen2022ost} explore the effect of data augmentation on generalization capability. They demonstrate that a detector trained on one GAN architecture can generalize surprisingly well to others GAN architectures with carefully designed data augmentation strategy. However, generalization ability is still a problem when diffusion models are engaged, because of the architectural difference between GANs and diffusion models. Zhong~\emph{et al.}~\cite{PatchCraft} develop a two-branch method based on the combination of patches, but this model exhibits poor robustness due to the introduction of rich texture patches. More recently, pretrained models are exploited to facilitate fake image detection~\cite{ojha2023fakedetect,tan2023learning,wang2023dire}. For example, ojha~\emph{et al.}~\cite{ojha2023fakedetect} use a pretrained CLIP-ViT~\cite{radford2021learning} to project the images to the latent space and then train a linear classifier in the latent space. However, as the variety of generative models increases, this method may not generalize well to all generators due to the frozen backbone network, which limits its capacity. Because most model parameters are fixed, it is unclear which factors contribute to the differentiation between fake and real images. Meanwhile, the inference speed is too slow due to the heavy model structure.

In this paper, we propose a novel method to tackle the abovementioned problems. We demonstrate that a single simple patch in the image can provide strong clue to differentiate between real and fake images. Specifically, we first  extract a single simple patch from the entire image, and then pass this patch through SRM filters~\cite{fridrich2012rich} to extract its noise fingerprints, based on which a binary classifier is employed to spot fake images. We name this method single simple patch (SSP) network, which can generalize well across different generators. Our method design, \emph{i.e.}, extracting the noise fingerprints of simple patch, is inspired by the camera capture process. Currently, the existing advanced generative models tend to focus on generating complex details in the generated images, while overlooking the consistency of simple contents between generated images and real images. Because real images are captured by cameras, various factors such as external interference and components within the camera itself introduce noise into the electronic signals during the photosensitive process. Intuitively, when a generative model produces a patch with pure color, it only needs to ensure that the pixels have similar colors within this patch. However, in the real images captured by cameras, it is almost impossible for the simple patches to have pure color, as they usually contain various types of noise. Besides, the complex patches in the images contain excessive redundant information, which may interfere with the decision-making process of detector. As a result, we choose simple patches to identify the fake images.

However, we observe that when the quality of generated images decreases (\emph{e.g.}, blur or compression artifacts), the detection performance is inclined to drop. Consequently, we insert an enhancement module and a perception module before SSP network to alleviate these artifacts. The perception module evaluates the probability of input patch being blurry, compressed, or intact. Such information is embedded into the enhancement module to produce the high-quality patch. Then the patch is input into the SSP network. Through extensive experiments, we find that the suppression of blur and compression has little impact on the noise fingerprints used by SSP network to distinguish between real and fake images. Therefore, the extra design of perception module and enhancement module could improve the detection capability on low-quality generated images. 

We conduct comprehensive experiments on GenImage Dataset~\cite{zhu2023gendet} and ForenSynths Dataset~\cite{wang2020cnn} to demonstrate the effectiveness of our method. Our contributions can be summarized as follows:
\begin{itemize}
    \item We propose a single simple patch (SSP) network, revealing that the noise fingerprints of simple patch serves as strong clue for AI-generated image detection. 
    \item We further improve our SSP network to effectively reduce the impact of poor image quality on detection difficulty. 
    \item Comprehensive experiments on two benchmark datasets show that our approach  can generalize well across different generators.
\end{itemize}

\section{RELATED WORKS}
\subsection{Generative Models}
In recent years, image generative models have garnered considerable attention. These models are capable of generating images that are very similar to real ones. Generative Adversarial Network~\cite{goodfellow2014generative} (GAN) is one early generative model proposed for image generation. GAN consists of a generator and a discriminator, in which the generator accounts for creating images while the discriminator accounts for determining whether the images are real or fake. These two components are learnt through adversarial training. Afterwards, many researchers have proposed the variant models of GAN. For example, BigGAN \cite{donahue2019large} uses a hierarchical generator architecture that allows control over image details and diversity at different levels, which facilitates the generation of more detailed and diverse images. Recently, diffusion model was introduced for image generation. It starts with random noise, which is gradually transformed using a diffusion process. The diffusion-based text-to-image generation models~\cite{radford2021learning,Nichol2021GLIDETP,rombach2022high} can be used to create images of high quality. 

\subsection{AI-Generated Image Detection}
As the images produced by generative models become increasingly realistic, it has become more challenging for humans to distinguish between real and fake images. In earlier works, researchers typically employed learning-based approaches to address this problem, treating it as a binary classification task. Zhang ~\emph{et al.}~\cite{zhang2019detecting} demonstrated that unique artifacts exist in GAN-generated images. These artifacts can be viewed periodically in the spectrum of images. Frank~\emph{et al.}~\cite{frank2020leveraging} trained the detector in the frequency domain to detect them. Wang~\emph{et al.}~\cite {wang2020cnn} proposed that with appropriate training data  and data augmentation, a classifier trained on one CNN generator can generalize well to other unseen generators. In \cite{tan2023learning}, the author used a pretrained CNN model to convert images to gradients, which are used to present the artifacts in GANs. 

However, after the emergence of diffusion models, these methods have shown poor generalization when applied to the images generated by diffusion models (DM). Ricker ~\emph{et al.}~\cite{ricker2022towards} found that DMs do not exhibit any obvious artifacts in the frequency domain and there is structural difference between images generated by DMs and GANs. Wang~\emph{et al.}~\cite{wang2023dire} used the error between an input image and its reconstruction by a pre-trained diffusion model to identify fake images, but this method cannot generalize well to GAN-based models. Ojha ~\emph{et al.}~\cite{ojha2023fakedetect} proposed UnivFD which utilizes the feature space of a frozen pretrained vision-language model followed by a linear classifier, but this model cannot be extended well with the emergence of new advanced generative models and the inference speed is much lower than other methods. Tan~\emph{et al.}~\cite{tan2023rethinking} proposed to use the local interdependence among image pixels caused by upsampling operators to detect the artifacts. Zhong~\emph{et al.}~\cite{PatchCraft} leverages two branches for rich and poor texture regions followed by carefully designed high-pass filters to classify the image. ChaiIn ~\emph{et al.}~\cite{chai2020makes} also proposed a method in patch level as ours. 
They utilizes truncation at different layers of the neural network (\emph{e.g.}, ResNet \cite{he2016deep}, Xception  \cite{Chollet_2017}) to obtain model predictions based on the collective information from the whole set of patches. This method can help localize the manipulated regions. 
However, it was proved in \cite{ojha2023fakedetect} that this method does not perform well in detecting AI-generated images. It struggles to effectively distinguish even the images generated by the same generator. Unlike their work, our method uses the information from the original images. We just use one single simple patch followed by typical filters \cite{fridrich2012rich}, which is capable of generalizing well across all the generators.

\section{METHOD}
\subsection{Overview}
\begin{figure*}[t]
\centering
\includegraphics[width=\linewidth]{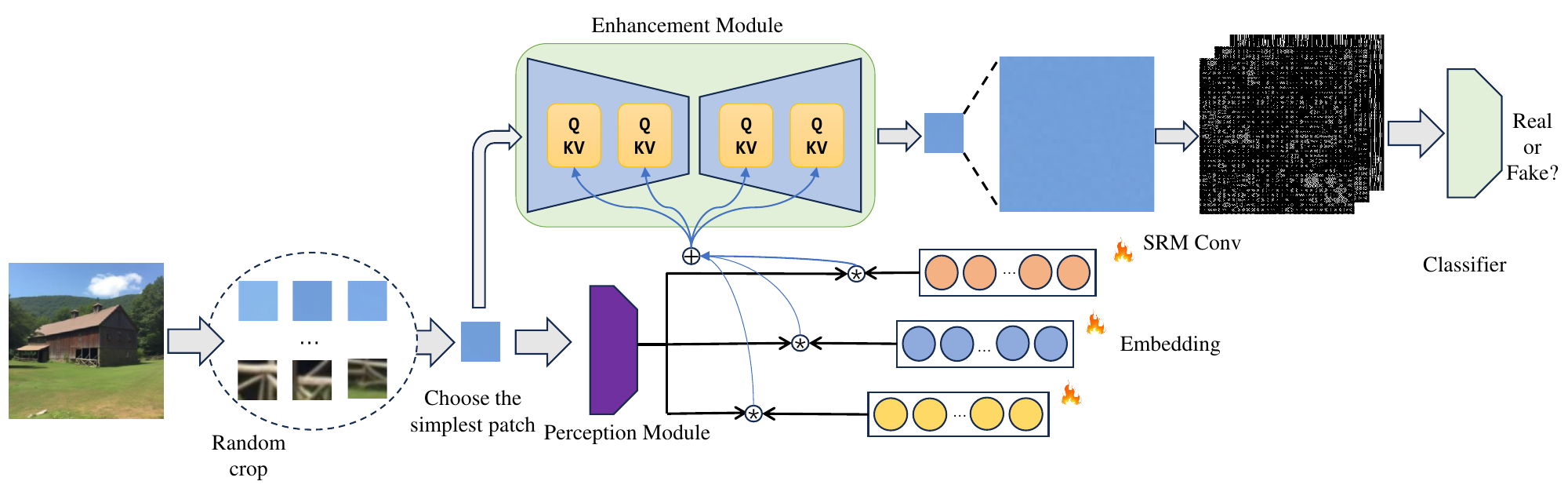} 
\caption {The overall architecture of our proposed method, which consists of three parts: enhancement module, perception module, and single simple patch (SSP) network. We first extract the simplest patch from the original image. Then, we use the enhancement module and perception module to get high-quality patch, which is sent to the SSP network.}
\label{Overview}
\end{figure*}

Given a test image, our task is to distinguish whether this image is real or not. The challenge lies in designing a detector that can generalize well across different generators. Inspired by \cite{chai2020makes,PatchCraft}, we attempt to solve this problem in the patch level. We observe that using only a single simple patch from the original image is enough to identify the fake image. The overview of the proposed model is illustrated in Figure~\ref{Overview}. In Section~\ref{method Sec:1}, we will first describe how to extract the simple patch and its noise pattern. In Section~\ref{Method Sec2}, we will introduce the enhancement module and perception module to address the issue of interference caused by low image quality.
\subsection{Single Simple Patch (SSP) Network}
\label{method Sec:1}
In this section, we introduce our single simple patch (SSP) network. Given an image $\bm{X}$ with spatial size $W\times H$, SSP predicts the output based on one simple patch. We first randomly split the image into many small patches with patch size $M \times M$. We treat the patch with the smallest texture diversity as the simplest patch. The texture diversity is measured by pixel fluctuation degree proposed in~\cite{PatchCraft}, which can be calculated as follows,
\begin{equation}\label{texture computation}
\begin{aligned}
\mathcal{L}_{ div } = \sum_{i = 1}^{M} \sum_{j = 1}^{M-1}(\left | x_{i,j} \!- \! x_{i,j+1} \right | )  \!+\! \sum_{i = 1}^{M-1} \sum_{j = 1}^{M}(\left | x_{i,j} \!-\! x_{i+1,j} \right | 
    \\ + \sum_{i = 1}^{M-1} \sum_{j = 1}^{M-1}(\left | x_{i,j} \!-\! x_{i+1,j+1} \right |  \!+\! \sum_{i = 1}^{M-1} \sum_{j = 1}^{M-1}(\left | x_{i+1,j} \!-\! x_{i,j+1} \right |
),
\end{aligned}
\end{equation}

\noindent in which $x_{i,j}$ is the $(i,j)$-th entry in the patch. $\mathcal{L}_{ div }$ consists of four components along the four directions (up-down, left-right, diagonal, and anti-diagonal). It calculates the absolute difference between adjacent entries in each direction and accumulates the differences, effectively estimating the complexity. 

Then, we resize the simplest patch with minimum $\mathcal{L}_{ div }$ to the size of original image, and use standard SRM (Steganalysis Rich Model)~\cite{fridrich2012rich} to extract its noise pattern, which records the high-frequency information by using $3$ high-pass filters. Finally, the noise pattern is delivered to ResNet50~\cite{he2016deep} binary classifier to distinguish real images from fake images. The output of the classifier represents the probability for the input image being a real image. SSP network employs a standard binary cross-entropy loss:
\begin{equation}
    \mathcal{L}_{cls} =  \hat{y} \log y^{\prime} + (1 - \hat{y})log(1-y^{\prime}),
\end{equation}
where $\hat{y}$ denotes the ground-truth real/fake label and $y^{\prime}$ denotes the predicted label of SSP. 

As mentioned in Section~\ref{sec:intro}, generative models, such as diffusion models, focus more on generating subtle details in the regions with rich textures to make the generated images more realistic, while the simple patches in generated images are largely overlooked. These simple patches actually contain lots of hidden noise caused by camera capture which can be used as important fingerprints. With these insights, our SSP network directly focuses on the simple patches, reducing the influence of irrelevant and cluttered information from the original image. Additionally, focusing on the simplest patch also reduces the computational cost, leading to a simple yet effective model.  

\subsection{Enhanced SSP Network}
\label{Method Sec2}
Our SSP network performs well in distinguishing between real and fake images in common situations. However, images may undergo a certain degree of degradation in the real world. In such scenarios, not only SSP but also other AI-generated detection methods would suffer from performance drop to some degree. Therefore, we design an enhancement module and a perception module to reduce the influence caused by interfering factors like blurring and compression. 

Given a patch, both the perception module and the enhancement module work together to transform a lower-quality patch into a higher-quality one. Our enhancement module has commonly used U-Net~\cite{ronneberger2015u} structure, which aims to enhance the patch quality if the input patch is blurry or compressed. 
Moreover, we design a perception module to provide guidance information (whether the input patch is blurry or compressed). The perception module is a lightweight three-class classifier (blurry, compressed, intact),  which consists of a convolution layer, batch normalization layer, ReLU activation function, a pooling layer and a fully connected layer.

Based on the prediction of perception module, the enhancement module can perform the corresponding task, which helps alleviate the burden of enhancement module. 
Specifically, if the input patch is blurry or compressed, enhancement module performs the deblurring and  decompression task respectively. If the patch is neither blurry nor compressed, enhancement module simply reconstructs the input patch. Corresponding to the abovementioned three tasks, we introduce three learnable task embeddings, which are combined based on the prediction of perception module and injected into the enhancement module as conditional information.

Formally, we use $F$ to denote the perception module, and use $\bm{w}'=[w'_1, w'_2, w'_3]$ to denote the predicted probability of input patch $\bm{x}$ being blurry, compressed, or intact, respectively. 
\begin{equation}
    \bm{w}'  = F(\bm{x}).\\
\end{equation} 
Then we get the normalized weights $\bar{\bm{w}}'=[\bar{w}'_1, \bar{w}'_2, \bar{w}'_3]$ by $\bar{\bm{w}}' =\frac{\bm{w}'}{||\bm{w}'||_1}$.
Based on the normalized weights, we combine three learnable task embeddings:
\begin{equation}
    \bm{h}^{fus}  = \bar{w}'_{1} \cdot \bm{h}^{blu} + \bar{w}'_{2} \cdot \bm{h}^{com} + \bar{w}'_{3} \cdot \bm{h}^{rec},
\end{equation}
in which $\bm{h}^{blu}, \bm{h}^{com}, \bm{h}^{rec}$ are the task embeddings for patch deblurring, decompression, and reconstruction, respectively. 

The fused embedding $\bm{h}^{fus}$  serves as the conditional information of enhancement module, aiming to assist the enhancement module in removing the interference factors. Recall that the enhancement module has U-Net structure. The fused embedding is injected into the encoding and decoding blocks in the enhancement module using cross-attention.  Specifically, the enhancement process can be written as follows,
\begin{equation}
    \bm{x}^{\prime}  = E(\bm{x}, \bm{h}^{fus}), \\
\end{equation}
in which $E$ denotes the enhancement module and $x^{\prime}$ denotes the output patch. 

During training, we apply Gaussian blur and JPEG compression to the original patch $\hat{\bm{x}}$ with certain probability and acquire $\bm{x}$ as the input patch of enhancement module. In the meanwhile, we can obtain the ground-truth perception labels $\hat{\bm{w}}=[\hat{w}_1, \hat{w}_2, \hat{w}_2]$. In particular,  if a patch is applied with Gaussian blur (\emph{resp.}, JPEG compression), we set $\hat{w}_1 =1$ (\emph{resp.}, $\hat{w}_2 =1$). If a patch remains the original patch, we set $\hat{w}_3 =1$. The perception module is trained with MSE loss. For the output patch of enhancement module, we also employ MSE loss, arriving at the following loss function:
\begin{equation}
    \mathcal{L}_{enh} = ||\bm{x}^{\prime}-\hat{\bm{x}}||^2 + ||\bm{w}' - \hat{\bm{w}}||^2.
\end{equation}
\noindent

\section{EXPERIMENTS}

\subsection{Dataset}
First, we conduct experiments on recently proposed GenImage dataset~\cite{zhu2023genimage} containing some advanced generators. GenImage uses the real images in ImageNet~\cite{deng2009imagenet} and their category labels to generate fake images. The fake images are generated by $8$ generators including Stable Diffusion V1.4~\cite{rombach2022high}, Stable Diffusion V1.5~\cite{rombach2022high}, GLIDE~\cite{Nichol2021GLIDETP}, VQDM~\cite{gu2022vector}, Wukong~\cite{Wukong}, BigGAN~\cite{donahue2019large}, ADM~\cite{dhariwal2021diffusion}, and Midjourney~\cite{Midjourney}. Each generator is associated with a subset of real images and fake images, which are split into training subset and test subset. Therefore, GenImage consists of 8 training subsets and 8 test subsets, corresponding to 8 generators. In total, GenImage contains 1,331,167 real images and 1,350,000 fake images. Following~\cite{zhu2023genimage}, we train the classifier on the training subset of one generator and evaluate the classifier on the test subset of another generator, leading to $64=8\times 8$ settings. We also evaluate the degraded image classification task on this dataset. 

To compare with previous works more thoroughly, we also conduct experiments on ForenSynths Dataset \cite{wang2020cnn} which consists of images across 20 categories. The training set is composed of images generated by ProGAN generator. As for the testing set, we test our model on 7 GAN-based generators (ProGAN~\cite{karras2018progressive}, CycleGAN~\cite{CycleGAN}, BigGAN~\cite{BigGAN}, StyleGAN~\cite{karras2019style}, GauGAN~\cite{GauGAN}, StarGAN~\cite{choi2018stargan}, STGAN~\cite{STGAN}), 6 diffusion-based generators 
 (DDPM~\cite{ho2020denoising}, PNDM~\cite{liu2022pseudo}, IDDPM~\cite{nichol2021improved}, Guided Diffusion Model~\cite{dhariwal2021diffusion}, GLIDE~\cite{Nichol2021GLIDETP}, LDM~\cite{rombach2022high}) and one autoregresive model (DALLE-E~\cite{ramesh2021zero}) from UniversalFake Dataset~\cite{ojha2023fakedetect}. 

\subsection{Evaluation Metrics} Following existing works for AI-generated image detection, we adopt accuracy (ACC) and mean average precision (mAP) as evaluation metrics. The threshold for computing accuracy is set to 0.5. 
\begin{table*}[thb]
	\centering
	\setlength{\tabcolsep}{0.4mm}
	\begin{tabular}{c|cccccccc|cc}
		\hline
		\multicolumn{1}{c|}{\multirow{2}{*}{Method}} & \multicolumn{8}{c|}{test subset} & \multicolumn{1}{c}{\multirow{2}{*}{Avg Acc. (\%)}} \\ 
		\cline{2-9}
		\multicolumn{1}{c|}{}  & Midjourney & SD V1.4  & SD V1.5 & ADM & GLIDE & Wukong & VQDM & BigGAN & \multicolumn{1}{c}{} \\ 
		\hline
		ResNet-50~\cite{he2016deep} & 54.9 & \textbf{99.9} & 99.7 & 53.5 & 61.9 & 98.2 & 56.6 & 52.0 & 72.1 \\ 
		DeiT-S~\cite{touvron2021training} & 55.6 & \textbf{99.9} & 99.8 & 49.8 & 58.1 & 98.9 & 56.9 & 53.5 & 71.6 \\ 
		Swin-T~\cite{liu2021swin} & 62.1 & 99.9 & 99.8 & 49.8 & 67.6 & \textbf{99.1} & 62.3 & 57.6 & 74.8 \\ 
		CNNSpot~\cite{wang2020cnn} & 52.8 & 96.3 & 95.9 & 50.1 & 39.8 & 78.6 & 53.4 & 46.8 & 64.2 \\ 
		Spec~\cite{zhang2019detecting} & 52.0 & 99.4 & 99.2 & 49.7 & 49.8 & 94.8 & 55.6 & 49.8 & 68.8 \\ 
		F3Net~\cite{qian2020thinking} & 50.1 & 99.9 & \textbf{99.9} & 49.9 & 50.0 & 99.9 & 49.9 & 49.9 & 68.7 \\ 
		GramNet~\cite{liu2020global} & 54.2 & 99.2 & 99.1 & 50.3 & 54.6 & 98.9 & 50.8 & 51.7 & 69.9 \\ 
            DIRE~\cite{wang2023dire} & 60.2 & 99.9 & 99.8 & 50.9 & 55.0 & 99.2 & 50.1 & 50.2 & 70.7\\
            UnivFD~\cite{ojha2023fakedetect} & 73.2 & 84.2 & 84.0 & 55.2 & 76.9 & 75.6 & 56.9 & 80.3 & 73.3 \\
            NPR~\cite{tan2023rethinking} & 64.9 & 99.2 & 99.1 & 49.8 & 52.3 & 98.3 & 50.0 & 49.7 & 71.6 \\
            
            PatchCraft~\cite{PatchCraft} & 79.0 & 89.5 & 89.3 & 77.3 & 78.4 & 89.3 & 83.7 & 72.4 & 82.3 \\ 
            ESSP (Ours) & \textbf{82.6} & 99.2 & 99.3 & \textbf{78.9} & \textbf{88.9} & 98.6 & \textbf{96.0} & \textbf{73.9}  & \textbf{90.6} \\
		\hline
	\end{tabular}
 	\caption{The results of different methods trained on SD V1.4 and evaluated on different test subsets. We train all the models using the same data augmentation, \emph{i.e.}, Gaussian blur and JPEG compression with the probability of 10\%. For each test subset, the best results are highlighted in boldface.}
	\label{tab:Performance Comparison}
\end{table*}

\begin{table*}[thb]
	\centering
	\setlength{\tabcolsep}{1.2mm}{
		\begin{tabular}{c|cccccccc|c}
			\hline
			\multicolumn{1}{c|}{\multirow{2}{*}{Training Subset}} & \multicolumn{8}{c|}{test subset}                                                 & \multirow{2}{*}{\begin{tabular}[c]{@{}c@{}}Avg \\      Acc.(\%)\end{tabular}} \\ \cline{2-9}
		\multicolumn{1}{c|}{}
                         & Midjourney & SD V1.4 & SD V1.5 & ADM & GLIDE  & Wukong & VQDM  & BigGAN & \\ \hline
		Midjourney   & \textbf{98.8}/\textbf{95.7} & 76.4/90.6 & 76.9/90.8 & 64.1/81.5 & 78.9/89.1 & 71.4/86.6 & 52.5/81.4 & 50.1/75.9 & 71.1/86.6\\
		SD V1.4      & 54.9/82.6 & \textbf{99.9}/\textbf{99.2} & 99.7/99.3 & 53.5/78.9 & 61.9/88.9 & 98.2/98.6 & 56.2/96.0 & 52.0/73.9  & 72.1/90.6 \\
	    SD V1.5      & 54.4/82.1 & 99.8/99.3 & \textbf{99.9}/\textbf{99.4} & 52.7/78.7 & 60.1/96.2 & 98.5/99.0 & 56.9/94.8 & 51.3/68.0 & 71.7/88.5 \\
            ADM          & 58.6/88.0 & 53.1/97.2 & 53.2/97.1 & \textbf{99.0}/\textbf{99.0} & 97.1/96.7 & 53.0/97.9 & 61.5/97.7 & 88.3/85.5 & 70.4/95.0 \\
		GLIDE        & 50.7/78.4 & 50.0/84.8 & 50.1/85.9 & 56.0/86.3 & \textbf{99.9}/\textbf{99.2} & 50.3/84.6 & 51.0/80.4 & 74.0/78.8 & 60.2/84.8 \\
		Wukong       & 54.5/85.2 & 99.7/99.3 & 99.6/99.3 & 51.4/90.0 & 58.3/96.7 & \textbf{99.9}/\textbf{99.1} & 58.7/98.3 & 50.9/87.2  & 71.6/94.6 \\
		VQDM         & 50.1/75.7 & 50.0/88.4 & 50.0/88.4 & 50.7/73.7 & 60.1/93.9 & 50.2/87.1 & \textbf{99.9}/\textbf{98.2} & 66.8/70.8 & 59.7/84.5 \\
		BigGAN       & 49.9/82.4 & 49.9/99.3 & 49.9/99.2 & 50.6/89.9 & 68.4/93.2 & 49.9/98.8 & 50.6/98.0 & \textbf{99.9}/\textbf{88.0} & 58.6/93.8 \\ 
            \hline
	\end{tabular}}
 	\caption{The results of ResNet50 and our ESSP method with different training and test subsets. In each slot, the left (\emph{resp.}, right) number is the result of ResNet50 (\emph{resp.}, ESSP). For each test subset, the best results are highlighted in boldface.}
	\label{tab:cross_generator}
\end{table*}
\begin{table*}[tb]
	\centering
	\setlength{\tabcolsep}{0.4mm}
	\begin{tabular}{c|cccccccc|cc}
		\hline
		\multicolumn{1}{c|}{\multirow{2}{*}{Method}} & \multicolumn{8}{c|}{test subset} & \multicolumn{1}{c}{\multirow{2}{*}{\begin{tabular}[c]{@{}c@{}}Avg \\ Acc. (\%)\end{tabular}}} \\ 
		\cline{2-9}
		\multicolumn{1}{c|}{}  & Midjourney & SD V1.4  & SD V1.5  & ADM     & GLIDE   & Wukong  & VQDM    & BigGAN  & \multicolumn{1}{c}{} \\ 
		\hline
		ResNet-50~\cite{he2016deep} & 59.0 & 72.3 & 72.4 & 59.7 & 73.1 & 71.4 & 60.9 & 66.6 & 66.9 \\ 
		DeiT-S~\cite{touvron2021training} & 60.7 & 74.2 & 74.2 & 59.5 & 71.1 & 73.1 & 61.7 & 66.3 & 67.6 \\ 
		Swin-T~\cite{liu2021swin} & 61.7 & 76.0 & 76.1 & 61.3 & 76.9 & 75.1 & 65.8 & 69.5 & 70.3 \\
		CNNSpot~\cite{wang2020cnn} & 58.2 & 70.3 & 70.2 & 57.0 & 57.1 & 67.7 & 56.7 & 56.6 & 61.7 \\ 
		Spec~\cite{zhang2019detecting} & 56.7 & 72.4 & 72.3 & 57.9 & 65.4 & 70.3 & 61.7 & 64.3 & 65.1 \\ 
		F3Net~\cite{qian2020thinking} & 55.1 & 73.1 & 73.1 & 66.5 & 57.8 & 72.3 & 62.1 & 56.5 & 64.6 \\ 
		GramNet~\cite{liu2020global} & 58.1 & 72.8 & 72.7 & 58.7 & 65.3 & 71.3 & 57.8 & 61.2 & 64.7 \\ 
            ESSP (Ours) & \textbf{83.7} & \textbf{94.7} & \textbf{84.8} & \textbf{88.0} & \textbf{94.2} & \textbf{94.0} & \textbf{93.1} & \textbf{78.5} & \textbf{88.8} \\
		\hline
	\end{tabular}
 	\caption{The results of different methods with different training and test subsets. For each test subset, we calculate the average of eight results obtained using eight training subsets. For each test subset, the best results are highlighted in boldface.}
	\label{tab: All Performance Comparison}
\end{table*}

\subsection{Implementation Details}
All the images are first resized to 256$\times$256, and then we randomly crop 64 patches from each image. We choose the simplest patch as the input. Concerning the classifier of our SSP network, we use ResNet-50~\cite{he2016deep} pretrained on ImageNet~\cite{deng2009imagenet}. During the training stage, we apply Adam optimizer with learning rate $10^{-4}$. The batch size is set to 64 and we train our model for 30 epochs in total. As for the data augmentation, we use Gaussian blur and JPEG compression with the probability of 10\%. The proposed method is implemented by PyTorch library. As for the hardware devices used for training, we use Intel(R) Xeon(R) Silver 4116 CPU, with 128GB memory and one NVIDIA GeForce RTX 3090 GPU.

\subsection{Experiments on GenImage Dataset} \label{sec:eesp GenImage}
We first conduct experiments on the recently proposed GenImage dataset which includes images generated by recent generators, so GenImage is more challenging than the ForenSynths dataset in Section~\ref{sec:eesp expr}. We
compare our ESSP method with ResNet50 across different training and test subsets. 
The results are summarized in Table~\ref{tab:cross_generator}, which shows that our method significantly outperforms ResNet50 when the training and test subsets are from different generators. 

In Table~\ref{tab: All Performance Comparison}, for each test subset, we average the eight results obtained using eight training subsets. Our method surpasses all the baseline methods by a large margin. To the best of our knowledge, our ESSP is currently the only method that performs well in such challenging cross-generator settings, further highlighting the importance of noise fingerprints of simple patches. 

Following \cite{zhu2023genimage}, we also compare our ESSP with other methods when training subset is stable diffusion V1.4~\cite{rombach2022high}. The results are recorded in Table~\ref{tab:Performance Comparison}. We can observe that most methods have relatively low accuracy, which indicates that GenImage is a challenging dataset for AI-generated image dataset. Our ESSP surpasses the state-of-the-art PatchCraft~\cite{PatchCraft} by 8.3\%, achieving an accuracy of 90.6\%. As for the method using large pretrained models, our ESSP achieves significant accuracy gain over DIRE~\cite{wang2023dire} and UnivFD~\cite{ojha2023fakedetect}, with improvements of 19.9\% and 17.3\%, respectively. This also demonstrates the limited capability of pretrained models in classifying recently emerging generative models. 

\subsection{Experiments on ForenSynths Dataset}
\label{sec:eesp expr}
In this section, we compare our ESSP with previous works on ForenSynths Dataset. The experimental setting is consistent with previous works. To evaluate the generalization ability, the detector is trained on the images generated by ProGAN~\cite{karras2018progressive} and evaluated on the images generated by various generators from different categories. Table~\ref{tab:Performance_Comparison_on_UniversalFakeDetect_acc} (\emph{resp.}, Table~\ref{tab:Performance_Comparison_on_UniversalFakeDetect_ap}) summarizes the accuracy (\emph{resp.}, mAP) of different methods on this dataset. The results show that our ESSP achieves competitive accuracy and mAP. It is worth noting that ESSP achieves an accuracy of 90.22\%, surpassing the state-of-the-art baselines. Although the mAP of ESSP is slightly worse than UnivFD~\cite{ojha2023fakedetect}, our ESSP  has lower computation cost and faster inference speed than UnivFD which uses pretrained CLIP-ViT.

\begin{table*}[tb]
    \begin{tabular}{l|cccccccc|c}
    \hline
    \multicolumn{1}{c|}{\multirow{2}{*}{Model setting}} & \multicolumn{8}{c|}{test subset} & \multicolumn{1}{c}{\multirow{2}{*}{\begin{tabular}[c]{@{}c@{}}Avg \\ Acc. (\%)\end{tabular}}} \\ 
    \cline{2-9}
    \multicolumn{1}{c|}{}  & Midjourney & SD V1.4  & SD V1.5  & ADM     & GLIDE   & Wukong  & VQDM    & BigGAN  & \multicolumn{1}{c}{} \\ 
    \hline
    (1) Patch size 8 $\times$ 8 & 69.0 & 92.0 & 91.6 & 84.2 & 88.7 & 91.1 & 85.2 & 76.3 & 85.0\\ 
    (2) Patch size 16$\times$16 & 81.9 & 97.3 & 97.0 & 93.2 & 96.6 & 96.9 & 94.7 & 85.2 & 93.0 \\ 
    (3) Patch size 64$\times$64 & 82.4 & 99.8 & 99.7 & 74.5 & 94.7 & 99.8 & 98.0 & 71.4 & 90.4 \\
    \hline
    (4) 2 patches & 80.7 & 99.4 & 99.4 & 87.5 & 99.4 & 99.2 & 98.7 & 82.1 & 93.3 \\ 
    (5) 3 patches & 81.1 & 99.3 & 99.4 & 87.1 & 99.2 & 99.2 & 97.9 & 81.2 & 93.0 \\
    (6) 4 patches & 82.2 & 99.8 & 99.6 & 96.4 & 99.5 & 99.6 & 98.2 & 82.6 & 94.7 \\
    \hline
    (7) complex patch & 86.9 & 99.8 & 99.8 & 68.0 & 53.0 & 99.9 & 80.4 & 51.5 & 80.7  \\
    (8) complex + simple & 70.8 & 98.5 & 98.4 & 73.8 & 81.7 & 97.4 & 81.5 & 49.2 & 82.1 \\
    \hline
    (9) SSP w/o SRM & 80.5 & 98.1 & 98.1 & 94.2 & 87.9 & 97.6 & 80.3 & 59.6 & 85.0 \\ 
    (10) complete SSP & 83.1 & 99.4 & 99.3 & 87.4 & 98.5 & 99.0 & 98.0 & 81.3 & \textbf{93.5} \\
    \bottomrule
    \end{tabular}
    \caption{Ablation studies on our SSP network. We train our model using stable diffusion V1.4 and test on other generators in GenImage dataset.}
    \label{tab:ssp ablation}
\end{table*}

\begin{table*}[t!]
{\small
    \centering
    \tabcolsep=0.1cm
    \resizebox{1.\linewidth}{!}{
        \begin{tabular}{cc cccccc c cc cc c ccc ccc c c}
            \toprule
            
            \multirow{2}{*}{\shortstack[c]{Detection\\method}}  & \multicolumn{7}{c}{Generative Adversarial Networks} & \multirow{2}{*}{DDPM} & \multirow{2}{*}{PNDM} & \multirow{2}{*}{IDDPM} &\multirow{2}{*}{Guided} & \multicolumn{3}{c}{LDM} & \multicolumn{3}{c}{Glide} & \multirow{2}{*}{DALL-E} & Total \\
            \cmidrule(lr){2-8}  \cmidrule(lr){13-15} \cmidrule(lr){16-18} \cmidrule(lr){20-20}

            & \shortstack[c]{Pro-\\GAN} & \shortstack[c]{Cycle-\\GAN} & \shortstack[c]{Big-\\GAN} & \shortstack[c]{Style-\\GAN} & \shortstack[c]{Gau-\\GAN} &  \shortstack[c]{Star-\\GAN}   &\shortstack[c]{ST-\\GAN}    &  &  &  &  &  \shortstack[c]{200\\steps} & \shortstack[c]{200\\w/ CFG} & \shortstack[c]{100\\steps} & \shortstack[c]{100\\27} & \shortstack[c]{50\\27} & \shortstack[c]{100\\10} & & \shortstack[c]{Avg.\\Acc.(\%)}
            \\ 
            \midrule	
            
            {\shortstack[c]{CNNDet}}~\cite{wang2020cnn}  & 99.99 & 85.20 & 70.20 & 85.70 & \underline{78.95} & 91.70 & 67.20 & 64.11 & 52.10 & 53.40  & 60.07 & 54.03 & 54.96 & 54.14 & 60.78 & 63.80 & 65.66 & 55.58 & 67.64 \\

            {\shortstack[c]{Patchfor}}~\cite{chai2020makes}   &  75.03 & 68.97 & 68.47 & 79.16 & 64.23 & 63.94 & 90.12 & 63.52 & 53.14 & 56.71 & 67.41 & 76.51 & 76.12& 75.77 & 74.81 & 73.28 & 68.52 & 67.91 & 70.20 \\ 
            
            

            {\shortstack[c]{Co-occurence}}~\cite{nataraj2019detecting} & 99.74 & 63.15 & 53.75 & 92.50 & 51.10 & 54.70 & 52.40 & 65.90 & 56.31 & 59.31  & 60.50 & 70.70 & 70.55 & 71.00 & 70.25 & 69.60 & 69.90 & 67.55 & 66.49\\

            \shortstack[c]{LGrad}~\cite{tan2023learning} & 99.83 & 86.94 & 86.63 & 91.08 & 78.46& 99.27 & 54.71 & 72.30 & 67.50 & 68.92 & 60.34 & 90.30 & \textbf{95.10} &\underline{93.61} & 86.13 & 90.30 & 83.24 & 67.60 & 81.80\\
            \shortstack[c]{Spec}~\cite{zhang2019detecting} & 49.90 & \textbf{99.90} & 50.50 & 49.90 & 50.30 & 99.70 & 50.60 & 53.30 & 58.60 & 50.40  & 50.90 & 50.40 & 50.40 & 50.30 & 51.70 & 51.40 & 50.40 & 50.00 & 56.58\\
            
            \shortstack[c]{DIRE}~\cite{wang2023dire} & \textbf{100.0} & 67.73 & 64.78 & 83.08 & 65.30 & \textbf{100.0} & 80.16 & 86.31 & \underline{86.36} & 72.37 & \underline{83.20}& 82.70& 84.05& 84.25& \underline{87.10} &\underline{90.80} & \underline{90.25} & 58.75& 81.51\\
            {\shortstack[c]{{PatchCraft}}}~\cite{PatchCraft}  & \textbf{100.0} & 70.17 & \textbf{95.8} & 92.77 & 71.58 & 99.97 & \underline{92.70} & 43.5 & 56.2 & 46.00 & 61.70 & 74.55 & 71.20 & 74.15 & 42.35 & 42.55 & 42.90 & 70.35 & 69.41 \\ 
            {\shortstack[c]{{UniFD}}}~\cite{ojha2023fakedetect}  & 99.90 & \underline{98.40} & \underline{94.65} & 84.95 & \textbf{99.40} & 96.75 & 83.35 & 71.50 & 86.15 & 73.35  & 69.65 & \textbf{94.40} & 74.00 & \textbf{95.00} & 78.50 & 79.05 & 77.90 & \underline{87.30} & \underline{85.78} \\ 
            {\shortstack[c]{{NPR}}}~\cite{tan2023rethinking}  & 99.75 & 71.90 & 90.10 & \underline{94.30} & 77.70 & \textbf{100.0} & 87.80 & \underline{88.00} & 82.30 & \underline{83.8}& 68.35 & 89.80 & 90.10 & 89.05 & 76.55 & 80.15 & 79.50 & 85.40 & 85.25  \\ \hline
            
            {\shortstack[c]{{ESSP}}}  & 97.05 & 83.25 & 68.65 & \textbf{96.05} & 57.85  & 95.00 & \textbf{96.45} & \textbf{95.00} & \textbf{96.45}  & \textbf{91.45} & \textbf{81.60} & \underline{93.40}& \underline{94.60} & 93.30 & \textbf{97.70} & \textbf{97.80} & \textbf{97.75} & \textbf{91.70}  & \textbf{90.22} \\
            
            \bottomrule
    \end{tabular}}
}
\caption{The accuracy results of different methods on ForenSynths dataset. For each test subset, the best results are highlighted in boldface and the second best results are underlined.}
\label{tab:Performance_Comparison_on_UniversalFakeDetect_acc}
\end{table*}

\begin{table*}[t!]
{\small
    \centering
    \tabcolsep=0.1cm
    \resizebox{1.\linewidth}{!}{
        \begin{tabular}{cc cccccc c cc cc c ccc ccc c c}
            \toprule
            \multirow{2}{*}{\shortstack[c]{Detection\\method}}  & \multicolumn{7}{c}{Generative Adversarial Networks} & \multirow{2}{*}{DDPM} & \multirow{2}{*}{PNDM} & \multirow{2}{*}{IDDPM} &\multirow{2}{*}{Guided} & \multicolumn{3}{c}{LDM} & \multicolumn{3}{c}{Glide} & \multirow{2}{*}{DALL-E} & Total \\
            \cmidrule(lr){2-8}  \cmidrule(lr){13-15} \cmidrule(lr){16-18} \cmidrule(lr){20-20}

            & \shortstack[c]{Pro-\\GAN} & \shortstack[c]{Cycle-\\GAN} & \shortstack[c]{Big-\\GAN} & \shortstack[c]{Style-\\GAN} & \shortstack[c]{Gau-\\GAN} &  \shortstack[c]{Star-\\GAN}   &\shortstack[c]{ST-\\GAN}    &  &  &  &  &  \shortstack[c]{200\\steps} & \shortstack[c]{200\\w/ CFG} & \shortstack[c]{100\\steps} & \shortstack[c]{100\\27} & \shortstack[c]{50\\27} & \shortstack[c]{100\\10} & & \shortstack[c]{mAP(\%)}
            
            \\ 
            
            \midrule
            
            {\shortstack[c]{CNNDet}}~\cite{wang2020cnn}  & \textbf{100.0} & 93.47 & 84.50 & \textbf{99.54} & 89.49 & 98.15 & 84.11 & 78.23 & 91.27 & 83.16  & 73.72 & 70.62 & 71.10 & 70.54 & 80.65 & 84.91 & 82.07 & 70.59 &  83.67\\
            
            {\shortstack[c]{Patchfor}}~\cite{chai2020makes}
            &  80.88 & 72.84 & 71.66 & 85.75 & 65.99 & 69.25 & 97.10 & 98.40 & 67.31 & 61.20  & 75.03 & 87.10 & 86.72 & 86.40 & 85.37  & 83.73 & 78.38 & 75.67 & 79.37 \\
            
            
            {\shortstack[c]{Co-occurence}}~\cite{nataraj2019detecting}   &  99.74 & 80.95 & 50.61 & 98.63 & 53.11 & 67.99 & 63.70 & 71.32 & 74.10 & 71.11  & 70.20 & 91.21 & 89.02 & 92.39 & 89.32 & 88.35 & 82.79 & 80.96 & 78.64\\
            
            {\shortstack[c]{{LGrad}}}~\cite{tan2023learning}  & \textbf{100.0} & 95.01 & 92.93 & 98.31 & \underline{95.43} & \textbf{100.0} & 67.20 & 74.81 & 71.60 & 76.40 & 69.60 & 91.52 & \textbf{99.31} & 94.20 & 89.50 & 96.20 & 89.70 & 78.90 & 87.81 \\	
            
            \shortstack[c]{Spec}~\cite{zhang2019detecting}  & 55.39 & \textbf{100.0} & 75.08 & 55.11 & 66.08 & \textbf{100.0} & 78.90 & 74.61 & 87.30 & 76.11& 57.72 & 77.72 & 77.25 & 76.47 & 68.58 & 64.58 & 61.92 & 67.77 & 73.36 \\
            
            \shortstack[c]{DIRE}~\cite{wang2023dire} & \textbf{100.0} & 76.73 & 72.8 & 97.06 & 68.44 & \textbf{100.0} & 83.20 & 89.11 & 87.45 & 87.23 & \textbf{94.29} & 95.17 & 95.43 & 95.77 & \underline{96.18} & \underline{97.31} & \underline{97.53} & 68.73 & 89.02 \\
            {\shortstack[c]{{PatchCraft}}}~\cite{PatchCraft}  & \textbf{100.0} & 85.26 & \textbf{99.42} & 98.96 & 81.33 & \textbf{100.0} & 98.24& 43.56 & 64.36 & 46.11 & 69.76 & 82.29 & 78.81 & 82.60 & 46.95 & 48.71 & 49.28 & 76.1 & 75.09  \\ 
            {\shortstack[c]{{UniFD}}}~\cite{ojha2023fakedetect}  & \textbf{100.0} & \underline{99.84} & \underline{99.23} & 97.43 & \textbf{99.98} & 99.53 & 92.15 & \underline{98.54} & \underline{99.23} & \underline{97.13} & 87.64 & \textbf{99.32} & 92.50 & \textbf{99.27} & 95.28 & 95.56 & 94.96 & \textbf{97.47} & \textbf{96.94} \\
            
            {\shortstack[c]{{NPR}}}~\cite{tan2023rethinking}  & \textbf{100.0} & 82.05 & {96.37} & 99.17 & 81.44 & \textbf{100.0} & \textbf{99.64} & 93.74 & 93.87 & 80.74 & 75.09 & 96.84 & 96.93 & 96.59 & 84.30 & 87.53 & 87.10 & 93.97 & 91.40 \\  \hline

            {\shortstack[c]{{ESSP}}}  & {99.53} & 92.26 & 74.75 & \underline{99.18} & 61.53 & 98.61 & \underline{99.52} & \textbf{98.72} & \textbf{99.52} & \textbf{97.44}  & \underline{90.90} & \underline{97.33} & \underline{97.79} & \underline{97.14} & \textbf{99.82} &\textbf{99.87} &\textbf{99.82} & \underline{95.96} & \underline{94.42}\\	
            \bottomrule
    \end{tabular}}
}

\caption{The mAP results of different methods on ForenSynths dataset. For each test subset, the best results are highlighted in boldface and the second best results are underlined.}
\label{tab:Performance_Comparison_on_UniversalFakeDetect_ap}
\end{table*}

\begin{table}[tbh]
\centering
    \begin{tabular}{c|ccc}
    \hline
    Method       & Blur & Compression \\
    \hline
    SSP      &  72.2 &  69.3 \\ 
    ESSP w/o perception & 78.6 & 70.2 \\
    ESSP     &  \textbf{81.4}  & \textbf{73.8} \\
    \hline
    \end{tabular}
    \caption{Ablation studies on our enhancement module and perception module. We train and test our model using stable diffusion V1.4}
    \label{tab: ESSP ablation}
\end{table}

\subsection{Ablation Study}
To validate the effectiveness of individual components in our method, we conduct ablation studies on GenImage dataset. We train our model on stable diffusion V1.4 subset and test it on the others. 
\begin{figure}[t]
\centering
\includegraphics[width=0.99\linewidth]{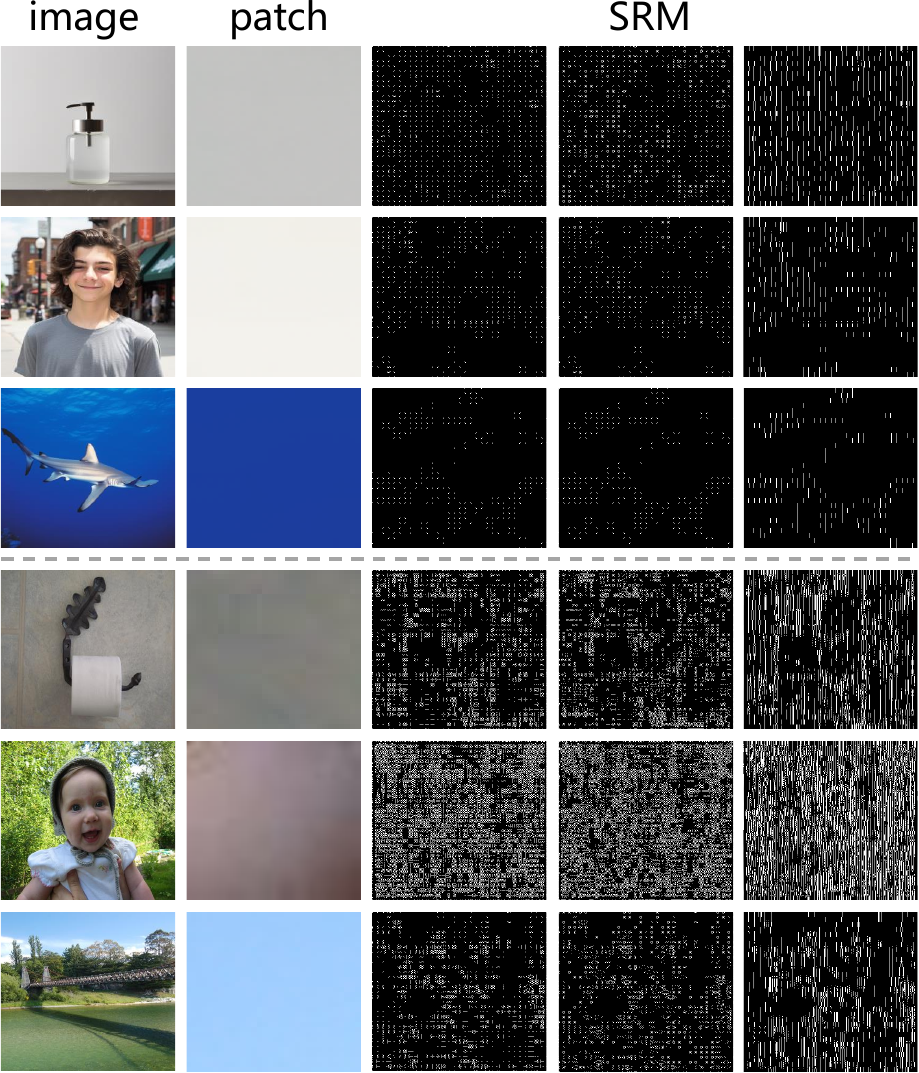}
\caption{In each row, from left to right, we show the image, the simplest patch, and SRM~\cite{fridrich2012rich}  outputs using 3 high-pass filters. The top three rows are fake images generated by Midjourney \cite{Midjourney}, while the bottom three rows are real images from ImageNet~\cite{deng2009imagenet}.}
\label{fig:SRM}
\end{figure}
\begin{figure*}[t]
\centering
\includegraphics[width=0.9\linewidth]{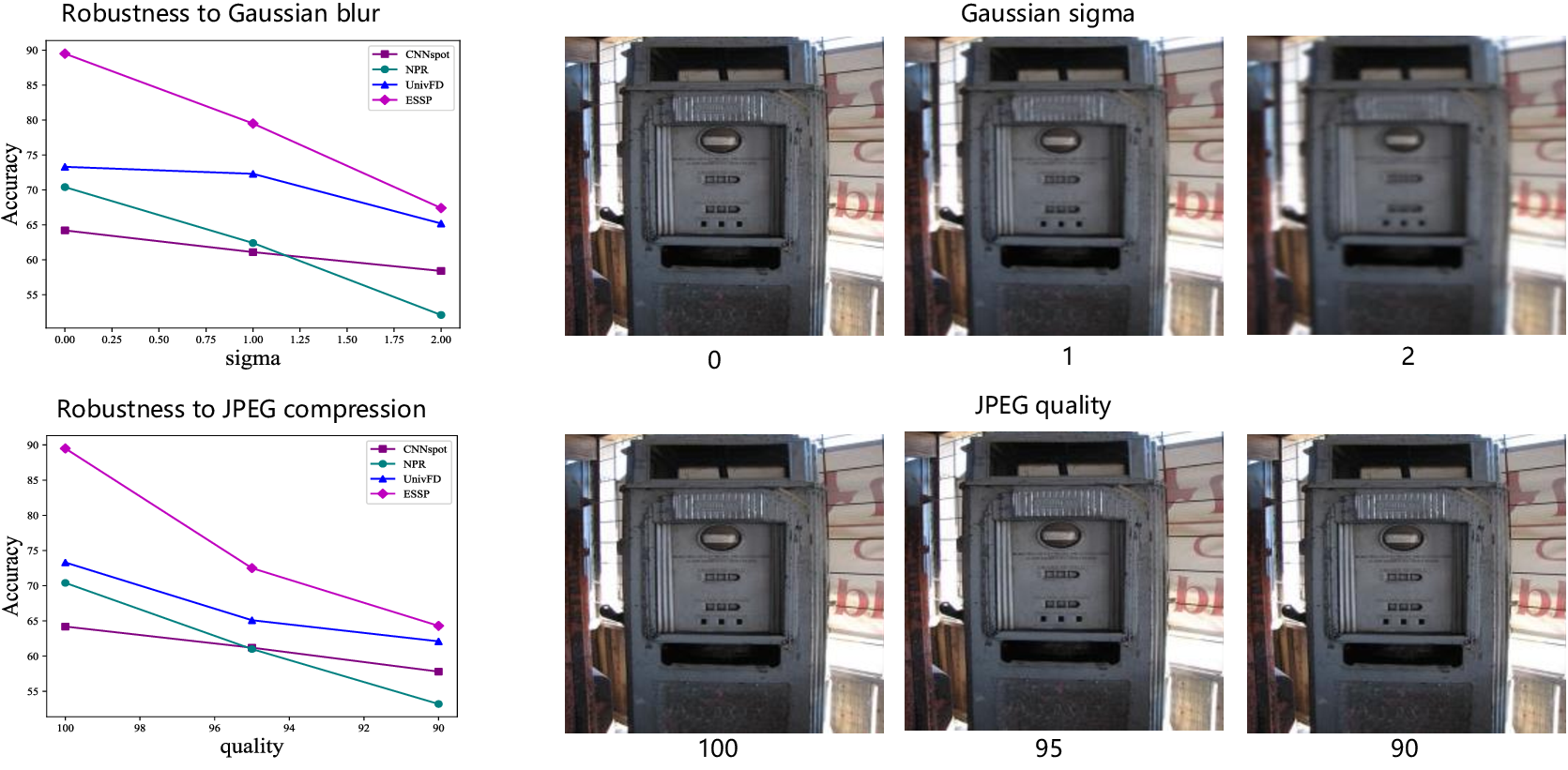}
\caption{Left: The robustness of different methods to blur and compression. Right: Example blurry or compressed images.}
\label{fig:robustness}
\end{figure*}

\subsubsection{Effectiveness of Simple Patch}
The key part of ESSP is the simplest patch. In Figure~\ref{fig:SRM}, we visualize several example real and fake images, along with their simplest patches and SRM outputs. It can be seen that the simplest patches have almost pure colors, but the noise fingerprints of real images and fake images are visibly different. 

We first investigate the impact of various patch attributes (\emph{e.g.}, patch size, the number of patches). The experiments are conducted based on SSP network, which is trained and tested without applying blur or compression. The results are listed in Table~\ref{tab:ssp ablation}, in which row 10 contains the results of default SSP network. 

Based on row 1-3, we observe that when the patch size is too large or too small, the performance is degraded. On one hand, overlarge patches include more complex information from the original image which may interfere with the judgment of model. On the other hand, the information of too small patches is inadequate for the model to make accurate judgment. 
Therefore, we ultimately choose the patch size of 32 for our model. 

As for the number of patches, previous works~\cite{PatchCraft,chai2020makes} propose to use a collection of multiple patches to determine the authenticity of an image. In row 4-6, we explore using more than one patch. In particular, we choose the top $k$ simplest patches and concatenate then along the channel dimension as the input of SSP network. When using 2 or 3 patches (row 4, 5), the obtained results are comparable with those using a single patch (row 10). When using 4 patches (row 6), the results become slightly better. 

In addition, we further conduct experiments using the complex patch extracted from the original image. We only use the most complex patch (row 7), or concatenate it with the simplest patch (row 8). The results drop sharply, compared with using the simplest patch. 

To verify the effectiveness of SRM filters, we remove SRM filters and directly send the patch to the classifier, leading to the results in row 9. The comparison between row 9 and row 10 justifies the necessity of SRM filters. 

\subsubsection{Effectiveness of Patch Enhancement} \label{sec:ablate_enhance}

We train and test our model on stable diffusion V1.4. Training images are processed by JPEG compression (QF $\sim$ Uniform[90, 100]) or Gaussian blur ($\sigma$ $\sim$ Uniform[0, 1]) with 10\% probability, which is the same as in Section~\ref{sec:eesp GenImage} and \ref{sec:eesp expr}. 
Different from Section~\ref{sec:eesp GenImage} and \ref{sec:eesp expr}, the test images are also blurred or compressed. 

The results are reported in Table~\ref{tab: ESSP ablation}. It can be seen that the performance of SSP is significantly degraded when the test images are blurred or compressed, possibly because the blur and compression interfere with the noise fingerprints used to distinguish between real and fake images. 
After employing an enhancement module to enhance the patch quality, the accuracy is significantly improved for blurry images and  compressed images. Furthermore, by incorporating a perception module and injecting the predicted information into the enhancement module, the accuracy is further improved. The accuracy reaches 81.4\% for blurry images and 73.8\% for compressed images. These results further prove the advantage of our perception module and enhancement module. 

\subsection{Robustness to Blur and Compression}
Following Section~\ref{sec:ablate_enhance}, we further explore the robustness of our method to blur and compression. 
We conduct experiments on GenImage dataset~\cite{zhu2023genimage}. 
We train the model on stable diffusion V1.4 generator and test it on eight generators in GenImage dataset. The eight results are averaged to achieve the final result. All models are trained with the same strategy as in Section~\ref{sec:ablate_enhance}.  For test images, we control the compression quality and blur degree. 

We compare our model with the state-of-the-art methods: UnivFD~\cite{ojha2023fakedetect}, NPR~\cite{tan2023rethinking}, CNNSpot~\cite{wang2020cnn}. The results of different methods are plotted in Figure~\ref{fig:robustness}.  It can be seen that as the image quality decreases, all models suffer from performance decline, which implies that the interfering factors  substantially raise the difficulty of AI-generated image detection task. Nevertheless, our model still outperforms the latest models. 

We also present the images that are processed by blurring or compression in Figure~\ref{fig:robustness}. When $\sigma$ is 2, the image quality is already very low. We believe that conducting forgery detection on excessively blurred images is not very meaningful. When JPEG compression quality is below 90, the accuracy of all methods is generally very low, so we set the limit of compression quality as 90 during experiments.

\subsection{Limitations}
Although our ESSP method can generalize well across a wide range of generators, there is still room for improvement in detecting  very low-quality images (\emph{e.g.}, compression quality is lower than 90). We would explore how to integrate simple patches with other information in the original image for better robustness.

\section{CONCLUSION}
In this paper, we have proposed a simple yet effective AI-generated image detection method. Our work focuses on the noise pattern of simple patch in the original images, which can effectively distinguish fake images from real images. Additionally, we have also introduced an enhancement module and a perception module to promote the robustness of our method on low-quality images. Extensive experiments verify the effectiveness of our ESSP for AI-generated image detection.

{
    \small
    \bibliographystyle{ieeenat_fullname}
    \bibliography{main.bbl}
}


\end{document}